\documentclass[letterpaper, 10 pt, conference]{ieeeconf}  % Comment this line out if you need a4paper

\IEEEoverridecommandlockouts                              % This command is only needed if 
\usepackage{graphics} % for pdf, bitmapped graphics files
\usepackage{epsfig} % for postscript graphics files
\usepackage{times} % assumes new font selection scheme installed
\usepackage{amsmath} % assumes amsmath package installed
\usepackage{amssymb}  % assumes amsmath package installed
\usepackage{amsfonts}
\usepackage[ruled, vlined]{algorithm2e}
\usepackage{graphicx}
\usepackage{epstopdf}
\usepackage{textcomp}
\usepackage{xcolor}

\usepackage{siunitx}
\newtheorem{defn}{Definition}
\newtheorem{thm}{Theorem}
\usepackage{multirow}

\newcommand{\sepsone}{\mathcal{S}_{\epsilon_1}}
\newcommand{\sepstwo}{\mathcal{S}_{\epsilon_2}}

\usepackage{mymacros}

\usepackage{lipsum}

\title{\LARGE \bf
Provably Safe Trajectory Optimization in the Presence of Uncertain Convex Obstacles
}

\author{Charles Dawson, Ashkan Jasour, Andreas Hofmann, and Brian Williams% <-this % stops a space
\thanks{The authors are with the Model-Based Embedded and Robotic Systems Lab at the Massachusetts Institute of Technology, Cambridge, MA 02139, USA {\tt\small \{cbd, jasour, hofma, williams\}@mit.edu}. This work was supported by Airbus SE.}%
}

\begin{document}

\maketitle
\thispagestyle{empty}
\pagestyle{empty}

%%%%%%%%%%%%%%%%%%%%%%%%%%%%%%%%%%%%%%%%%%%%%%%%%%%%%%%%%%%%%%%%%%%%%%%%%%%%%%%%
\begin{abstract}

Real-world environments are inherently uncertain, and to operate safely in these environments robots must be able to plan around this uncertainty. In the context of motion planning, we desire systems that can maintain an acceptable level of safety as the robot moves, even when the exact locations of nearby obstacles are not known. In this paper, we solve this \textit{chance-constrained motion planning} problem using a sequential convex optimization framework. To constrain the risk of collision incurred by planned movements, we employ geometric objects called $\epsilon$-shadows to compute upper bounds on the risk of collision between the robot and uncertain obstacles. We use these $\epsilon$-shadow-based estimates as constraints in a nonlinear trajectory optimization problem, which we then solve by iteratively linearizing the non-convex risk constraints. This sequential optimization approach quickly finds trajectories that accomplish the desired motion while maintaining a user-specified limit on collision risk. Our method can be applied to robots and environments with arbitrary convex geometry; even in complex environments, it runs in less than a second and provides provable guarantees on the safety of planned trajectories, enabling fast, reactive, and safe robot motion in realistic environments.

\end{abstract}

%%%%%%%%%%%%%%%%%%%%%%%%%%%%%%%%%%%%%%%%%%%%%%%%%%%%%%%%%%%%%%%%%%%%%%%%%%%%%%%%
\section{Introduction}

In an ideal world, robots could trust the maps they use to navigate; however, anyone who has deployed a robotic system in the real world will know how difficult it is to obtain a perfectly accurate map. Localization uncertainty, sensor noise, human unpredictability, and other factors all ensure that few models survive contact with the real world. Robots must instead plan using uncertain models, optimizing performance while limiting the amount of risk incurred in accomplishing their goals. In the context of motion planning, this challenge takes the form of optimizing a planned trajectory while limiting the probability of collision with nearby obstacles (e.g. a human with unpredictable future actions).

Motion planning in the absence of uncertainty is a well-studied problem. Approaches such as sampling-based planning (RRT), optimization techniques (TrajOpt \cite{schulmanFindingLocallyOptimal2013}), and combined sampling-optimization planners (as in \cite{daiImprovingTrajectoryOptimization2018}) have demonstrated impressive performance in a wide range of applications. There have been several attempts to apply similar techniques in the uncertain case \cite{blackmoreChanceConstrainedOptimalPath2011,axelrodProvablySafeRobot2018,ludersChanceConstrainedRRT2010,vandenbergLQGMPOptimizedPath2011,sunSafeMotionPlanning2016,daiChanceConstrainedMotion2018}, including modified RRT algorithms \cite{ludersChanceConstrainedRRT2010,axelrodProvablySafeRobot2018} and outer-loop optimization approaches \cite{daiChanceConstrainedMotion2018,vandenbergLQGMPOptimizedPath2011}.

Unfortunately, many uncertainty-aware planners lack support for complex geometry in the robot or environment, limiting their usefulness in many applications (e.g. industrial pick-and-place or home robotics). When the robot can be approximated as a point and environment as a system of linear inequalities, collision risk can be estimated analytically, allowing for efficient path planning under uncertainty \cite{blackmoreChanceConstrainedOptimalPath2011,sunSafeMotionPlanning2016,masahiroonoIterativeRiskAllocation2008}. In these simplified environments, techniques such as mixed-integer or disjunctive mathematical programming have been applied to great effect \cite{blackmoreChanceConstrainedOptimalPath2011}. In contrast, approaches dealing with more complex geometry (i.e. unions of convex shapes) cannot rely on analytical solutions. Instead, they typically rely on computationally expensive sampling- or numerical integration-based techniques for estimating collision risk, yielding estimates that are not easily differentiable (making optimization difficult) and not guaranteed to be accurate (limiting the safety of such approaches) \cite{parkEfficientProbabilisticCollision2017,daiChanceConstrainedMotion2018}.

One notable approach to chance-constrained motion planning in the literature is that of Dai et al., which employs an iterative optimization approach to satisfying chance constraints and is notable for supporting complex robot and environment geometry \cite{daiImprovingTrajectoryOptimization2018}. This approach uses a quadrature sampling approach to estimate collision risk stemming from uncertainty in the robot's state, but this reliance on sampling leads to slow performance and a lack of safety guarantees.
 
\begin{figure}[t]
  \centering
  \includegraphics[width=\linewidth]{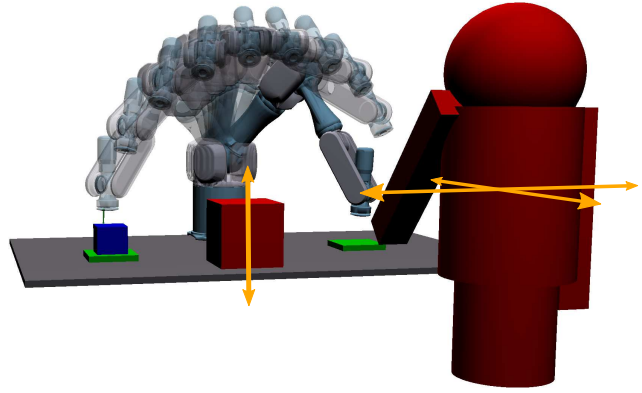}
  \caption{An example of a chance-constrained trajectory for a pick-and-place task, where the robot navigates around uncertain obstacles (red, with arrows indicating directions in which the obstacles' locations are most uncertain).}
  \label{fig:optimized_trajectory}
\end{figure}

In applications such as home robotics or collaborative manufacturing, uncertainty in the state of the environment often dominates any uncertainty in the robot's own state (simply because joint encoders are much more accurate than visual object-detection systems). Furthermore, when developing robots to operate near humans, we seek to provide guarantees on system safety. As a result, there is a need for algorithms that can quickly plan motions through uncertain environments with rich geometry (as in Fig.~\ref{fig:optimized_trajectory}) while providing strong guarantees on safety. 

\subsection{Contributions}

This paper makes two main contributions. First, we extend our previous work on risk estimation \cite{dawsonFastCertificationCollision2020} to compute the gradient of collision risk with respect to robot state, through the use of geometric objects called \textit{$\epsilon$-shadows} to certify upper bounds on collision risk. These risk estimates and gradients can be computed quickly even when the robot and environment have complex geometry; moreover, the $\epsilon$-shadow formulation provides theoretical guarantees that the true risk of collision will never exceed the certified bound. We focus our approach on the case where the robot and environment can be represented as collections of convex shapes and where the positions of obstacles in the environment are subject to additive Gaussian uncertainty.

For our second contribution, we use these risk certificates to optimize robot trajectories subject to a constraint on the risk of collision over the entire trajectory, employing a framework we call sequential convex optimization with risk allocation (SCORA). Using this framework, which we implement on top of existing sequential optimization solvers, we demonstrate path planning with bounded collision risk in simulated environments with non-trivial geometry. This optimization approach not only runs quickly ($<\SI{0.5}{s}$ in our experiments) but provides formal guarantees on the safety of optimized trajectories.

\subsection{Notation}

In the following discussion, we use script symbols (e.g. $\mathcal{X}$, $\mathcal{O}$) to denote subsets of $\R^n$, such as the set of points occupied by one link of a robot or the set of points occupied by an obstacle. In the case when obstacles are subject to additive uncertainty in position, we denote the \textit{nominal} geometry of the obstacle, located at the obstacle's expected position, as $O \subset \R^n$. Formally, the set of points $\mathcal{O}$ actually containing the uncertain obstacle is related to $O$ by $\mathcal{O} = \set{x + d : x \in O}$, where the uncertain translation $d \sim \mathcal{N}(0, \Sigma)$ is a zero-mean multivariate Gaussian random variable with covariance $\Sigma$. We will restrict our analysis to convex obstacle geometry, since non-convex geometries can be approximated as convex decompositions \cite{mamouSimpleEfficientApproach3D2009}.

\section{Risk certificates and gradient calculation}\label{sec:risk_estimate}

This section reviews our method for estimating the collision risk between a robot and its environment and extends this approach to provide the gradient of these risk estimates as well. We begin by introducing the concept of an $\epsilon$-shadow: a geometric object that is guaranteed to contain an uncertain obstacle with some probability. In particular, as in our previous work on risk estimation \cite{dawsonFastCertificationCollision2020}, we follow Axelrod, Kaelbling, and Lozano-P\'erez in considering only \textit{maximal} $\epsilon$-shadows, which we define below \cite{axelrodProvablySafeRobot2018}.

\begin{defn}\label{max_eshadow_defn}
    {\normalfont (maximal $\epsilon$-shadow)} A set $\mathcal{S} \subseteq \R^n$ is a maximal $\epsilon$-shadow of an uncertain obstacle $\mathcal{O}$ if the probability $P(\mathcal{O} \subseteq \mathcal{S}) = 1-\epsilon$.
\end{defn}

Intuitively, a maximal $\epsilon$-shadow is a shape that completely contains an uncertain obstacle with some specified probability. If there exists an $\epsilon$-shadow of an obstacle $\mathcal{O}$ that does not contact the robot, then the risk that the robot collides with that obstacle is guaranteed to be no greater than $\epsilon$. Using these $\epsilon$-shadows, the problem of finding an upper bound on collision risk reduces to the problem of finding a large $\epsilon$-shadow for each obstacle (equivalently, a small upper bound $\epsilon$) that does not intersect with the robot.

How one constructs such shadows varies based on how obstacle uncertainty is modeled. In their work on this subject, Axelrod, Kaelbling, and Lozano-P\'erez model uncertain obstacles as the intersection of linear inequalities with normally-distributed coefficients \cite{axelrodProvablySafeRobot2018}. This model naturally captures the uncertainty of point-cloud obstacles (e.g. LIDAR data), but for obstacles detected by other means (e.g. a visual pose estimation system) it is more natural to model the object as a known shape $O$ with position subject to additive Gaussian noise. By choosing this uncertainty model, we can guarantee that the corresponding $\epsilon$-shadows will be convex as long as the underlying obstacles are convex, enabling efficient collision checking (the same is not true for Axelrod, Kaelbling, and Lozano-P\'erez's formulation). For a proof of this fact, the reader is referred to our previous work \cite{dawsonFastCertificationCollision2020}. Here, we will review our approach for finding large $\epsilon$-shadows that do not intersect the robot, then present an extension that also provides the gradient of collision risk with respect to robot state.

\subsection{Estimating risk using $\epsilon$-shadows}

\begin{figure}[tbp]
    \centering
    \includegraphics[width=0.9\linewidth]{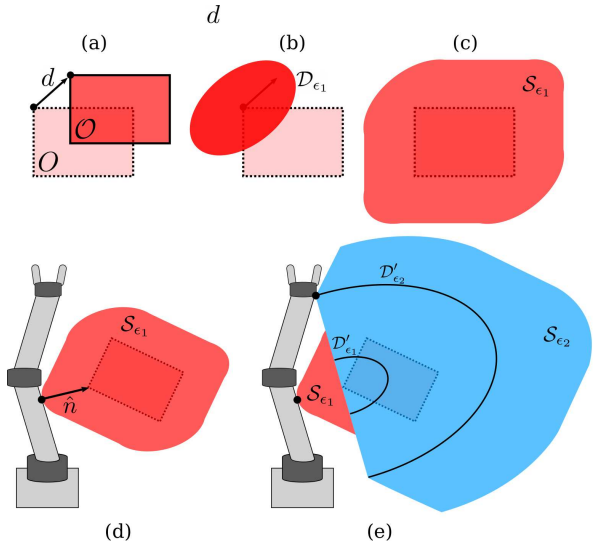}
    \caption{An illustration of the $\epsilon$-shadow approach to finding certifiable collision risk estimates. (a) The uncertain obstacle is translated from its nominal location by some uncertain vector. (b) We construct an ellipsoid to contain that translation vector with probability $1-\epsilon_1$. (c) We sum that ellipsoid with the nominal geometry to produce the $\epsilon_1$-shadow, guaranteed by Theorem~\ref{one-shot-proof} to contain the obstacle with probability $1-\epsilon_1$. (d) We find the largest such shadow that does not intersect the robot, but this estimate is conservative. (e) By expanding the $\epsilon_1$-shadow in the $\hat{n}$ direction away from the robot, as described in Algorithm~\ref{alg:two-shot}, we find a larger shadow $\sepsone \cap \sepstwo$ that yields a less conservative estimate $(\epsilon_1 + \epsilon_2)/2$.}
    \label{fig:shadow-recap}
\end{figure}

Recall that the uncertain obstacle $\mathcal{O}$ is related to the nominal geometry $O$ by $\mathcal{O} = \set{x + d : x \in O}$, as shown in Fig.~\ref{fig:shadow-recap}a, where $d\sim\mathcal{N}(0, \Sigma)$ is a zero-mean Gaussian random variable. By the properties of the multivariate Gaussian distribution, if we define the ellipsoid
$\mathcal{D}_{\epsilon_1} = \set{d: d^T\Sigma^{-1}d \leq \phi^{-1}(1-\epsilon_1)}$ (as in Fig.~\ref{fig:shadow-recap}b),
where $\phi^{-1}$ is the inverse cumulative distribution function (CDF) of the chi-squared distribution with $n$ degrees of freedom, then $P(d \in \mathcal{D}_{\epsilon_1}) = 1-\epsilon_1$ \cite{axelrodProvablySafeRobot2018}.

Next, we define the set $\mathcal{S}_{\epsilon_1}$ as the Minkowski sum of the nominal obstacle shape and the ellipsoid $\mathcal{D}_{\epsilon_1}$, as illustrated in Fig.~\ref{fig:shadow-recap}c, so that
\begin{equation}
  \mathcal{S}_{\epsilon_1} = \set{x + d : x \in O, d \in \mathcal{D}_{\epsilon_1}} \label{eq:one-shot-shadow}
\end{equation}

\begin{thm}\label{one-shot-proof}
$\mathcal{S}_{\epsilon_1}$ is convex and a maximal $\epsilon_1$-shadow of $O$.
\end{thm}
\begin{proofsk}
  It is helpful to think of $\mathcal{D}_{\epsilon_1}$ (centered at some point in $O$) as an $\epsilon_1$-shadow of that point in $\mathcal{O}$. By taking the Minkowski sum of $O$ and $\mathcal{D}_{\epsilon_1}$, we create a shape that contains every point in $\mathcal{O}$ with probability $1-\epsilon_1$, and thus contains $\mathcal{O}$ with probability $1-\epsilon_1$. Since the Minkowski sum of two convex shapes is itself convex, $\mathcal{S}_{\epsilon_1}$ is convex iff the nominal geometry $O$ is convex ($\mathcal{D}_{\epsilon_1}$ is always convex).
  % A more detailed proof:
  %
  % We observe that the event that $\mathcal{S}_{\epsilon_1}$ contains $\mathcal{O}$ is equivalent to the event that for all $y \in \mathcal{O}$, there exists both an $x\in O$ and a $d \in \mathcal{D}_{\epsilon_1}$ such that $x+d=y$. By construction, every every $y \in \mathcal{O}$ equals some $x + d$ for $x \in O$ and $d\sim\mathcal{N}(0, \Sigma)$, so we reduce the event $\mathcal{O} \subseteq \mathcal{S}_{\epsilon_1}$ to the event that $d \in \mathcal{D}_{\epsilon_1}; d\sim\mathcal{N}(0, \Sigma)$. Thus,
  % \begin{equation}
  %   P(\mathcal{O} \subseteq \mathcal{S}_{\epsilon_1}) = P(d \in \mathcal{D}_{\epsilon_1}; d\sim\mathcal{N}(0, \Sigma)) = 1-\epsilon_1
  % \end{equation}
  %
  % Thus, $\mathcal{S}_{\epsilon_1}$ is a maximal $\epsilon$-shadow of $\mathcal{O}$; furthermore, since the Minkowski sum of two convex sets is convex, and all ellipsoids are convex and $O$ is convex by assumption, $\mathcal{S}_{\epsilon_1}$ is convex as well.
  A detailed proof can be found in \cite{dawsonFastCertificationCollision2020}.
\end{proofsk}

To find the largest such $\mathcal{S}_{\epsilon_1}$ that does not contact the robot (correspondingly, the smallest risk bound $\epsilon_1$), we can employ a simple bisection line search; however, $\epsilon$-shadows of this form (shown in Fig.~\ref{fig:shadow-recap}d) tend to be very conservative. Because the robot is considered at risk of collision whenever the obstacle protrudes beyond its $\epsilon$-shadow, even if it protrudes \textit{away} from the robot, this simple ellipsoid-sum approach has a high false-positive rate.

To find a less conservative estimate that still certifies an upper bound on collision risk, we can exploit the geometric structure of the problem. As shown in Fig.~\ref{fig:shadow-recap}d, there is often free space extending away from the robot where the $\epsilon$-shadow can expand without intersecting the robot. We can take advantage of this free space by preferentially expanding the $\epsilon$-shadow in that direction until a second intersection occurs, as shown in Fig.~\ref{fig:shadow-recap}e.

To achieve this preferential expansion, we begin with a single ellipsoid-sum $\epsilon$-shadow $\sepsone$ that contacts the robot at one point and augment it with a second shadow constructed from a half-ellipsoid, as illustrated in Fig.~\ref{fig:shadow-recap}e.
If $\hat{n}$ is the contact normal between $\sepsone$ and the robot, we define the half-ellipsoid
% \begin{equation}
  $\mathcal{D}'_{\epsilon_2} = \set{d : d^T\Sigma^{-1}d \leq \phi^{-1}(1-\epsilon_2), \hat{n}^Td \geq 0}$
% \end{equation}
and corresponding convex shadow $\sepstwo$ (extending away from the robot in the $\hat{n}$ direction) as the Minkowski sum
\begin{equation}
  \sepstwo = \set{x + d : x\in O, d\in \mathcal{D}'_{\epsilon_2}} \label{eq:two-shot-shadow}
\end{equation}

We can find this expanded shadow using a second line search, which we warm-start with the results of the first search, as illustrated in Algorithm~\ref{alg:two-shot} \cite{dawsonFastCertificationCollision2020}. Because of this warm-start, Algorithm~\ref{alg:two-shot} will always find some $\epsilon_2 \leq \epsilon_1$ ($\epsilon_2 = \epsilon_1$ only when there is no surrounding free space). By taking the union of this second shadow $\sepstwo$ with the original $\sepsone$, we obtain the larger shadow $\mathcal{S}_{\epsilon'} = \sepsone \cup \sepstwo$ that provides a tighter bound $\epsilon' = (\epsilon_1 + \epsilon_2)/2$ on collision risk by capturing more of the collision-free workspace.

\begin{algorithm}[tb]
\SetAlgoLined
\SetKwInOut{Input}{Input}
\Input{Obstacle $\mathcal{O}$, covariance $\Sigma$, and tolerance $\epsilon_{tol}$}
\KwResult{Risk estimate $\epsilon$}
 Conduct bisection search to find smallest $\epsilon_1 \in (0, 1)$ to precision $\epsilon_{tol}$ such that the $\epsilon_1$-shadow of $\mathcal{O}$ constructed using Eq.~\eqref{eq:one-shot-shadow} does not intersect robot\;
 $\hat{n} \gets$ unit vector from the robot into the obstacle at the point of collision\;
 
 Conduct bisection search to find smallest $\epsilon_2 \in (0, \epsilon_1)$ to precision $\epsilon_{tol}$ such that the $\epsilon_2$-shadow of $\mathcal{O}$ constructed using Eq.~\eqref{eq:two-shot-shadow} does not intersect robot\;

 $\epsilon \gets (\epsilon_1 + \epsilon_2)/2$
 \caption{Multiple-expansion method for certifying bounds on robot-obstacle collision risk.}\label{alg:two-shot}
\end{algorithm}

\begin{thm}
  $\mathcal{S}_{\epsilon'}$ is a maximal $\epsilon'$-shadow of $\mathcal{O}$.
\end{thm}
\begin{proofsk}
  By the symmetry of the multivariate Gaussian (under reflection about the origin), we see that $\mathcal{D}'_{\epsilon_2}$ is a maximal $\epsilon_2/2$-shadow for the uncertain translation $d\sim\mathcal{N}(0, \Sigma)$. By the same reasoning used in Theorem~\ref{one-shot-proof}, the sum $\mathcal{S}_{\epsilon_2}$ is a maximal $\epsilon_2/2$-shadow of $\mathcal{O}$.
  %
  % First we remark, following the reasoning from our proof of Theorem~\ref{one-shot-proof}, that
  % \begin{equation}
  %   P(\mathcal{O} \subseteq \sepstwo) = P(d \in \mathcal{D}'_{\epsilon_2}; d \sim\mathcal{N}(0, \Sigma))
  % \end{equation}
  % Since the zero-mean multivariate Gaussian distribution is symmetric under reflection about any hyperplane containing the origin, the half-ellipsoid $\mathcal{D}'_{\epsilon_2}$ contains exactly half the probability mass as the corresponding full ellipsoid, so
  % \begin{equation}
  %   P(\mathcal{O} \subseteq \sepstwo) = P(d \in \mathcal{D}'_{\epsilon_2}; d \sim\mathcal{N}(0, \Sigma)) = \frac{\epsilon_2}{2}
  % \end{equation}

  % Next, we note that the event $\mathcal{O} \subseteq \mathcal{S}' = \sepsone \cup \sepstwo$ reduces to the event that $d \in \mathcal{D}_{\epsilon_1} \vee d \in \mathcal{D}'_{\epsilon_2}$. We can compute this probability directly:
  Since $\mathcal{O}$ is offset from $O$ by $d\sim\mathcal{N}(0, \Sigma)$, we see that
  \begin{align}
    P(\mathcal{O} \subseteq \mathcal{S}_{\epsilon'}) &= P(d \in \mathcal{D}_{\epsilon_1}) + P(d \in \mathcal{D}'_{\epsilon_1}) \nonumber \\ & \quad - P(d \in \mathcal{D}_{\epsilon_1} \cap \mathcal{D}'_{\epsilon_2}) \\
    &= \epsilon_1 + \frac{\epsilon_1}{2} - P(d \in \mathcal{D}_{\epsilon_1} \cap \mathcal{D}'_{\epsilon_2})
  \end{align}

  Since Algorithm~\ref{alg:two-shot} ensures $\epsilon_2 \leq \epsilon_1$, we can simplify the intersection $\mathcal{D}_{\epsilon_1} \cap \mathcal{D}'_{\epsilon_2}$ to
  % \begin{equation}
    $\mathcal{D}'_{\epsilon_1} = \set{d : d^T\Sigma^{-1}d \leq \phi^{-1}(1-\epsilon_1), \hat{n}^Td \geq 0}.$
  % \end{equation}
  %
  Again, by the symmetry of the Gaussian distribution, we see that $P(d \in \mathcal{D}'_{\epsilon_1}) = \epsilon_1/2$, so $P(\mathcal{O} \subseteq \mathcal{S}_{\epsilon'}) = (\epsilon_1 + \epsilon_2)/2$
  %
  % \begin{equation}
  %   P(\mathcal{O} \subseteq \mathcal{S}_{\epsilon'}) = \frac{\epsilon_1 + \epsilon_2}{2}
  % \end{equation}
\end{proofsk}

An important benefit of this risk estimation approach (compared to that used by Axelrod, Kaelbling, and Lozano-P\'erez) is that both $\sepsone$ and $\sepstwo$ are convex shapes, allowing fast collision checking between the $\epsilon$-shadows and the robot. Moreover, we can check for collision using implicit Minkowski sums (represented as support vector mappings), avoiding the computational cost of constructing set-wise sums. Support vector mappings represent convex shapes as functions taking directions to points on the shape furthest in that direction; they are commonly used in computational geometry algorithms \cite{gilbertFastProcedureComputing1988}. For a more detailed proof of the correctness of this approach, or for details on implementation and performance, the reader is referred to \cite{dawsonFastCertificationCollision2020}; our intention here is to extend this method to compute the collision risk gradient and develop a trajectory optimization framework, not to re-derive the basic algorithm.

\subsection{Derivation of risk gradient}

A key advantage of our technique is that these risk estimates are differentiable with respect to robot state, unlike other estimation methods in the literature (e.g. sampling \cite{daiChanceConstrainedMotion2018}). The existence of this derivative will allow us to efficiently incorporate collision risk as a constraint in a nonlinear optimization framework.

\begin{figure}[tpb]
  \centering
  \includegraphics[width=0.5\linewidth]{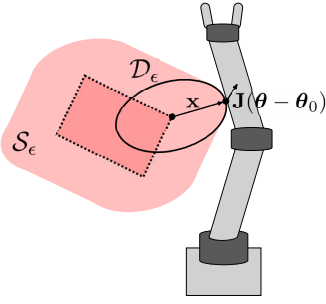}
  \caption{To derive the gradient, we isolate the component ellipsoid that makes contact with the robot.}
  \label{fig:gradient}
\end{figure}

The basis for our derivation of the gradient is summarized in Fig.~\ref{fig:gradient}. Once we have found the largest $\epsilon$-shadow (correspondingly, smallest $\epsilon$) that lies tangent to the robot, we can isolate the ellipsoid $\mathcal{D} = \set{d : d^T\Sigma^{-1}d \leq \phi^{-1}(1-\epsilon)}$ that makes contact with the robot, centered at some point on $O$. For simplicity, we will consider just one shadow $\mathcal{S}_\epsilon$, but our approach can be extended to the combined shadow $\mathcal{S}_{\epsilon'} = \sepsone \cup \sepstwo$, since the gradient of the combined risk estimate $\epsilon' = (\epsilon_1 + \epsilon_2)/2$ is simply the linear combination of the gradients of $\epsilon_1$ and $\epsilon_2$.

Since the size of this isolated ellipsoid $\mathcal{D}_{\epsilon}$ implicitly sets the risk estimate $\epsilon$, we can examine how this ellipsoid changes with small perturbations to the robot state. We assume that these small joint angle perturbations $\delta\v{\theta} \ll \v{\theta}_0$ do not change the location of $\mathcal{D}_{\epsilon}$, for example by causing its center to move to another point on $O$.

Let $\v{x}$ be the vector from the center of $\mathcal{D}_{\epsilon}$ to the point of contact. This vector can be calculated easily, as it is simply the support vector of $\mathcal{D}_{\epsilon}$ in the $-\hat{n}$ direction. Due to the definition of $\mathcal{D}_{\epsilon}$,
\begin{equation}
  \v{x}^T \Sigma^{-1} \v{x} = \phi^{-1}(1-\epsilon),
\end{equation}
so we can express the risk estimate in terms of $\v{x}$, as
\begin{equation}
  \epsilon = 1 - \phi(\v{x}^T \Sigma^{-1} \v{x}).
\end{equation}
This expression is differentiable, so we derive the gradient in terms of the chi-squared probability density function (PDF) with $n$ degrees of freedom, $\chi^2_n$,
\begin{equation}
  \nabla_{\v{x}}\epsilon = - \chi^2_n(\v{x}^T \Sigma^{-1} \v{x}) \pn{2 \v{x}^T \Sigma^{-1}} \label{eq:nablax_eps}
\end{equation}
By linearizing the robot's pose around its current joint state $\v{\theta}_0$ and computing the Jacobian $\v{J}$ at the point of contact, we can express the change in $\v{x}$ in terms of a small change in joint state $\delta\v{\theta}$ as
\begin{equation}
  \delta\v{x} = \v{J}\delta\v{\theta} \label{eq:dxdtheta}
\end{equation}
Combining Eqs.~\eqref{eq:nablax_eps} and~\eqref{eq:dxdtheta} yields the gradient of estimated risk with respect to the robot's joint state:
\begin{equation}
  \nabla_{\v{\theta}}\epsilon = - \chi^2_n(\v{x}^T \Sigma^{-1} \v{x}) \pn{2 \v{x}^T \Sigma^{-1}}\v{J}
\end{equation}

Using this gradient, we can linearize the risk of collision with each obstacle for a given robot configuration $\v{\theta}_0$ (where $\epsilon_0$ is the risk estimated for that configuration):
\begin{equation}
  \epsilon(\v{\theta}) \approx \epsilon_0 - \chi^2_n(\v{x}^T \Sigma^{-1} \v{x}) \pn{2 \v{x}^T \Sigma^{-1}}\v{J} \pn{\v{\theta} - \v{\theta}_0} \label{eq:eps_lin}
\end{equation}

Since $\v{x}$ can be calculated quickly using support-vector geometry and $\v{J}$ can be queried from any external kinematics engine, this gradient can be constructed with relatively little computational effort. With this linearization in hand, we can employ sequential convex optimization to find near-locally optimal trajectories that maintain desired risk levels.

\section{Sequential convex trajectory optimization}

Optimizing the trajectory of a robot moving around obstacles is a challenging non-convex optimization problem, even in the absence of uncertainty. Furthermore, complex robot and environment geometries make it difficult to apply traditional linear or quadratic programming methods, many of which use a point-robot approximation \cite{blackmoreChanceConstrainedOptimalPath2011}. Instead, deterministic trajectory optimizers that support complex geometry, such as TrajOpt \cite{schulmanFindingLocallyOptimal2013}, solve this problem using sequential convex optimization (SCO).

At a basic level, SCO, seeks to solve the non-convex optimization problem
\begin{align}
  &\text{minimize}\quad f(\v{x}) \label{eq:sco} \\
  &\text{subject to}\quad g_i(\v{x}) \leq 0,\quad i=1,2,\ldots,n_{ineq} \\
  &\phantom{subject to}\quad h_j(\v{x}) = 0,\quad j=1,2,\ldots,n_{eq}
\end{align}
where $f$, $g_i$, and $h_j$ are (possibly non-convex) objective, inequality constraint, and equality constraint functions, respectively, and $\v{x}$ is a vector of decision variables (typically joint angles $\v{\theta}_t$ at each of $T$ timesteps). Since $f$, $g_i$, and $h_j$ can be non-convex, SCO repeatedly constructs and optimizes a convex approximation of Problem~\eqref{eq:sco} until the true non-convex constraints are satisfied and the solution converges to a local optimum of $f$. Some SCO solvers (including TrajOpt) also dualize the constraints by incorporating a cost penalty for constraint violations, which improves performance when initialized with an infeasible solution \cite{schulmanFindingLocallyOptimal2013}.

In our work, we extend the SCO approach to consider constraints on the probability of collision in addition to deterministic collision-avoidance constraints. In the absence of uncertainty, one avoids collision by constraining the signed distance $\text{sd}_O(\v{\theta}_t)$ between the robot and each obstacle $O$ to be greater than some fixed safety margin at each timestep \cite{schulmanFindingLocallyOptimal2013}. When the obstacles' locations are uncertain, we need to consider not only the risk of collision at each timestep but also how risk accumulates over the entire trajectory. In practice, we would like to limit the risk incurred by the robot over the entire plan rather than merely limiting risk at each timestep (this type of constraint is referred to as a joint chance constraint). We enforce the joint chance constraint by adding the additional constraint \eqref{eq:joint_risk}, which allows the optimizer to intelligently \textit{allocate} risk across timesteps, taking more risk at some points and less at others in order to satisfy the joint risk threshold. In previous work on chance-constrained optimization, this allocation has been shown to improve the performance of risk-aware systems, as it allows the system to spend and save risk at different times as needed to achieve its objective  \cite{blackmoreChanceConstrainedOptimalPath2011,masahiroonoIterativeRiskAllocation2008}.

To formalize this approach, which we call SCORA (sequential convex optimization with risk allocation), we denote the estimated risk of collision with obstacle $O$ in state $\v{\theta}_t$ as $\epsilon_O(\v\theta_t)$, which we incorporate into the optimization problem:
\begin{align}
  &\text{minimize}\quad \sum_{t=1}^{T-1} \left\lVert \v{\theta}_t - \v{\theta}_{t-1} \right\rVert^2 \quad \text{subject to} \label{eq:optprob} \\
  % & \\
  &\quad \text{sd}_O(\v{\theta}_t) \geq d_{margin}, \quad t=0,\ldots,T-1;\ \forall O \label{eq:sd}\\
  &\quad \sum_{O} \epsilon_O(\v{\theta}_t) \leq \delta_t, \qquad\qquad t=0,\ldots,T-1 \label{eq:indiv_risk}\\
  &\quad \sum_{t=0}^{T-1} \delta_t \leq \Delta \label{eq:joint_risk}
\end{align}

Our decision variables in this problem are the joint angles at each timestep $\v{\theta}_t$ and the risk allocations $\delta_t$. We assume trajectory tracking is accomplished using a low-level controller and do not consider dynamics. Since the risk of collision at each timestep is bounded by $\delta_t$, the risk of collision during the entire plan cannot exceed $\sum_{t=0}^{T-1} \delta_t$. Thus, constraints~\eqref{eq:indiv_risk} and~\eqref{eq:joint_risk} ensure that the joint chance constraint is met while providing the freedom to take on more or less risk as needed to achieve good performance. We retain the constraint on signed distance in our formulation because the collision risk estimates saturate at $1$ when the robot is in contact with the nominal obstacle geometry, so this constraint is needed to penalize contact and escape risk-saturated configurations.

Although the objective~\eqref{eq:optprob} and joint chance constraint~\eqref{eq:joint_risk} are convex functions of the decision variables $\v{\theta}_t$ and $\delta_t$, the functions $\text{sd}(\v{\theta})$ and $\epsilon(\v{\theta})$ are non-convex in general, so we need to construct a convex approximation of Problem~\eqref{eq:optprob} about the current solution $\v{\theta}_{t, 0}$ and $\delta_t$. We can linearize the individual chance constraints \eqref{eq:indiv_risk} according to Eq.~\eqref{eq:eps_lin} and the signed distance function \eqref{eq:sd} according to \cite{schulmanFindingLocallyOptimal2013}, yielding the convex approximation:
\begin{align}
  &\text{minimize}\quad \sum_{t=1}^{T-1} \left\lVert \v{\theta}_t - \v{\theta}_{t-1} \right\rVert^2 \quad \text{subject to}\\
  % & \nonumber\\
  &\quad \text{sd}(\v{\theta}_{t, 0}) + \hat{n}^T\v{J}\pn{\v{\theta}_t - \v{\theta}_{t, 0}} \geq d_{margin}, \nonumber \\ &\qquad\qquad\qquad\qquad\qquad t=0,\ldots,T-1; \ \forall O\\
  &\quad \sum_{O} \left.\left[\epsilon_O(\v{\theta}_{t, 0}) - \chi^2_n(\v{x}^T \Sigma^{-1} \v{x}) \pn{2 \v{x}^T \Sigma^{-1}}\v{J} \pn{\v{\theta} - \v{\theta}_{t, 0}}\right]\right|_O \nonumber\\ &\qquad\qquad\qquad\  \leq \delta_t, \quad t=0,\ldots,T-1 \\
  &\quad \sum_{t=0}^{T-1} \delta_t \leq \Delta
\end{align}

This approximation is a quadratic program that can be solved quickly using an off-the-shelf convex optimizer. Of course, additional linear (or convexifiable) constraints can be added to this formulation if desired (e.g. to enforce joint angle limits).

\section{Results}

We implement the risk estimation algorithm presented in Section~\ref{sec:risk_estimate} using the C++ Bullet collision checking library. A detailed analysis of the performance of this algorithm in environments of varying complexity is presented in \cite{dawsonFastCertificationCollision2020}; in summary, Algorithm~\ref{alg:two-shot} provides strong upper bounds on collision risk and runs in less than \SI{200}{\micro s} in test environments. Here, we extend this implementation to provide the gradient of the risk estimate along with the estimate itself.

\begin{figure}[tpb]
  \centering
  \includegraphics[width=\linewidth]{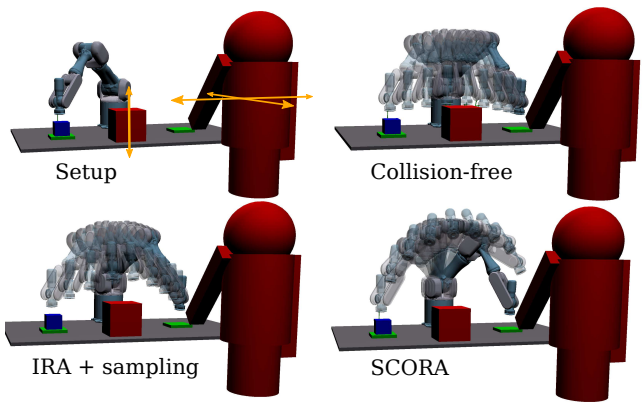}
  \caption{Optimized trajectories in our example scenario. The orange arrows indicate directions in which obstacle locations are uncertain. Our SCORA approach yields the most conservative trajectory, but it is the only approach that satisfies the $1\%$ collision risk constraint.}
  \label{fig:trajectories}
\end{figure}

\begin{figure}[tpb]
  \centering
  \includegraphics[width=\linewidth]{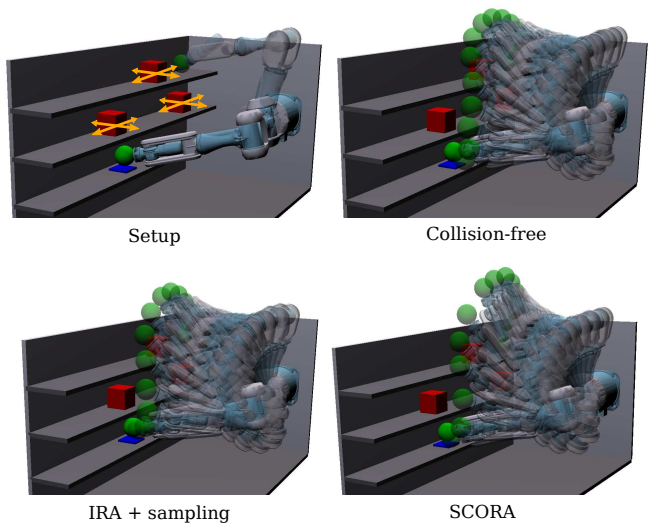}
  \caption{Optimized trajectories with a $10\%$ collision risk constraint in a pick-and-place task. Our SCORA approach not only satisfies the chance constraint but also runs much faster than the IRA algorithm.}
  \label{fig:trajectories2}
\end{figure}

We implement our SCORA optimization framework using TrajOpt's built-in SCO solver, to which we add our risk estimation, gradient, and allocation methods \cite{schulmanFindingLocallyOptimal2013}. All experiments were run on an Intel i9-7960X CPU. We tested our approach in the scenarios shown in Figs.~\ref{fig:trajectories} and~\ref{fig:trajectories2} (where orange arrows indicate uncertainty in obstacle location, with standard deviations ranging between 7-\SI{30}{cm}). We set a joint chance constraint of $1\%$ and $10\%$ for the scenarios in Figs.~\ref{fig:trajectories} and~\ref{fig:trajectories2}, respectively.

To benchmark our approach, we provide two comparisons. The first is with a trajectory optimized without any chance constraints. This risk-blind comparison provides a baseline for both the length of the optimal collision-free path and the time needed to find that path, allowing us to quantify the cost of imposing chance constraints in later examples.

The second comparison is with a trajectory optimized using the iterative risk allocation (IRA) method presented presented by Dai et al. in \cite{daiChanceConstrainedMotion2018}. This approach uses sampling to approximate the risk of collision at each timestep and repeatedly re-solves a deterministic optimization with safety margins $d_{margin}$ adjusted to penalize regions of high collision risk. The IRA algorithm presented in \cite{daiChanceConstrainedMotion2018} focuses on uncertainty in robot state, but we adapt this approach to consider obstacle uncertainty instead. This method provides a means for comparing SCORA against a previously-published chance-constrained trajectory optimization algorithm.

Our results are shown in Table~\ref{tab:results}. The collision-free optimization was seeded with straight lines in joint space, while IRA and SCORA were seeded with the output of the collision-free optimization.

\begin{table}[h]
\caption{Comparison of trajectory optimization algorithms}\label{tab:results}
\begin{center}
\begin{tabular}{c|c||c|c|c}
& Algorithm & Runtime$^1$ & Path length$^1$ & Collision \\
& & (s) & (rad) & risk$^2$ \\
\hline
Fig.~\ref{fig:trajectories} & Risk-blind & 0.005 & 2.35 & 59.16\% \\
$\Delta = 1\%$ & IRA & 0.480 & 2.65 & 20.85\% \\
$T=10$ & SCORA & 0.195 & 4.56 & 0.48\% \\
\hline
Fig.~\ref{fig:trajectories2} & Risk-blind & 0.154 & 7.21 & 18.87\% \\
$\Delta = 10\%$ & IRA & 2.202 & 7.22 & 9.22\% \\
$T=17$ & SCORA & 0.358 & 7.72 & 1.65\% \\
\multicolumn{4}{l}{$^1$ Averaged over $1,000$ trials.} \\
\multicolumn{4}{l}{$^2$ Averaged over $100,000$ trials.}
\end{tabular}
\end{center}
\end{table}

Since neither scenario is a ``piano-mover'' problem where very few feasible paths exist, we see that the risk-blind optimization quickly finds a nominally collision-free trajectory in both cases. Unsurprisingly, this trajectory is not robust to uncertainty in obstacle position. The IRA approach is somewhat more robust, but its reliance on sampling to estimate collision risk means that even though the algorithm converges, it often significantly underestimates the true risk of collision. In particular, the sampling strategy it employs cannot accurately measure low probabilities. As a result, IRA successfully achieves the $10\%$ risk bound but fails to achieve the $1\%$ bound.

In contrast, the SCORA approach proposed here not only satisfies the chance constraint in both cases (due to the formal guarantees of the $\epsilon$-shadow method) but also converges $2.5$-$6$ times more quickly than IRA. Our approach yields longer, more conservative trajectories, in part due to the conservatism of the $\epsilon$-shadow bound, but this performance-safety trade-off is typical in risk-aware systems, as the robot can improve its performance by taking on more risk.

\section{Conclusions}

In this paper, we develop a sequential convex optimization approach for solving the chance-constrained motion planning problem. To quickly estimate the probability of collision between the robot and uncertain obstacles, we use $\epsilon$-shadows to simultaneously certify an upper bound on collision risk and derive the gradient of that  risk with respect to robot state. This gradient allows us to construct a convex approximation of the chance-constrained trajectory optimization problem, which we solve efficiently using sequential convex optimization with risk allocation (SCORA).

We demonstrate our approach in simulation, yielding planning times under \SI{0.5}{s} ($2.5$-$6$ times faster than previously-published approaches) while ensuring that the optimized trajectory respects the user-specified risk bound. Because $\epsilon$-shadow risk certificates are guaranteed never to underestimate the true risk of collision, our approach produces provably safe motion plans even in the presence of obstacles with rich geometry and uncertain location.

\addtolength{\textheight}{-12cm}   % This command serves to balance the column lengths

\bibliographystyle{IEEEtran}
\bibliography{IEEEabrv,ccTrajOptPaper}

\end{document}